\pdfoutput=1
\documentclass[5pt]{article}

\usepackage[letterpaper]{geometry}
\usepackage{spconf,amsmath,epsfig}
\usepackage{enumitem}

\usepackage{verbatim}
\usepackage{graphicx}
\usepackage{color}
\usepackage{subcaption} 
\usepackage{etoolbox}
\usepackage{enumitem}

\usepackage[toc,page]{appendix}

\patchcmd{\thebibliography}
{\settowidth}
{\setlength{\parsep}{0pt}\setlength{\itemsep}{0pt plus 0pt}\settowidth}
{}{}

\usepackage{fancyhdr}
\begin{document}\sloppy

\def\x{{\mathbf x}}
\def\L{{\cal L}}

\title{Low-Latency Human Action Recognition with Weighted Multi-Region Convolutional Neural Network}
%
\name{Yunfeng Wang$^{\star}$, Wengang Zhou$^{\star}$, Qilin Zhang$^{\dagger}$, Xiaotian Zhu$^{\star}$, Houqiang Li$^{\star}$\thanks{This work is the extended version of \cite{wang2018weighted}. This work was supported in part to Dr. Houqiang Li by 973 Program under contract No. 2015CB351803 and NSFC under contract No. 61390514, and in part to Dr. Wengang Zhou by NSFC under contract No. 61472378 and No. 61632019, the Fundamental Research Funds for the Central Universities, and Young Elite Scientists Sponsorship Program By CAST (2016QNRC001).}}
\address{$^{\star}$University of Science and Technology of China, Hefei, Anhui, China\\
$^{\dagger}$HERE Technologies, Chicago, Illinois, USA
}
%
%

\maketitle

\begin{abstract}
  Spatio-temporal contexts are crucial in understanding human actions in videos. Recent state-of-the-art Convolutional Neural Network (ConvNet) based action recognition systems frequently involve 3D spatio-temporal ConvNet filters, chunking videos into fixed length clips and Long Short Term Memory (LSTM) networks. Such architectures are designed to take advantage of both short term and long term temporal contexts, but also requires the accumulation of a predefined number of video frames (e.g., to construct video clips for 3D ConvNet filters, to generate enough inputs for LSTMs). For applications that require low-latency online predictions of fast-changing action scenes, a new action recognition system is proposed in this paper. Termed ``Weighted Multi-Region Convolutional Neural Network" (WMR ConvNet), the proposed system is LSTM-free, and is based on 2D ConvNet that does not require the accumulation of video frames for 3D ConvNet filtering. Unlike early 2D ConvNets that are based purely on RGB frames and optical flow frames, the WMR ConvNet is designed to simultaneously capture multiple spatial and short term temporal cues (\emph{e.g.}, human poses, occurrences of objects in the background) with both the primary region (foreground) and secondary regions (mostly background). On both the UCF101 and HMDB51 datasets, the proposed WMR ConvNet achieves the state-of-the-art performance among competing low-latency algorithms. Furthermore, WMR ConvNet even outperforms the 3D ConvNet based C3D algorithm that requires video frame accumulation. In an ablation study with the optical flow ConvNet stream removed, the ablated WMR ConvNet nevertheless outperforms competing algorithms.
  \end{abstract}
  \begin{keywords}
  Action Recognition, Low-Latency, Online Prediction, Multi-region, ConvNet
  \end{keywords}
  \section{Introduction}
  \label{sec:intro}
  Generally, there are two types of action recognition methods, one based on the handcrafted conventional features and the other based on deep neural networks \cite{huang2018video,wang2018enhanced,zang2018attention,duan2018joint,lv2018video}. Among the former type, the iDT~\cite{wang2013action} achieves the state-of-the-art performance but it is also excessively computational expensive,  due to its requirements of dense video trajectories. Thanks to the advancements in deep learning \cite{ran2017convolutional,ran2017hyperspectral}, recent Convolutional Neural Network (ConvNet/CNN) based action recognition systems frequently involve 3D spatio-temporal ConvNet filters \cite{Tran_2015_ICCV,carreira2017quo} and chunking videos into fixed length clips. Such architectures are capable of exploiting even long term temporal contexts, but they require the accumulation of a predefined number of frames (e.g., to construct video clips for 3D ConvNet filters). For applications that require low-latency online predictions, alternative design of action recognition system need to be developed.

  \begin{figure}[t]
  \centering
  \includegraphics[scale=0.35]{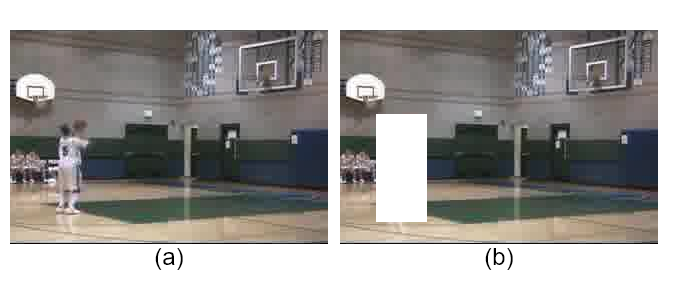}\\
  \caption{Background offers helpful hints in determining human actions. (a) Sample basketball video frames. (b) Masked video frame. Even with the basketball player censored, the surrounding scenes (basketball court, which could be covered by multiple ``secondary regions" in the proposed WMR ConvNet) still strongly suggest the action of ``playing basketball".}
  \label{bask_vs}
  \end{figure}

  Alternatively, multi-stream ConvNet (typically with an optical flow stream\footnote{Multi-stream ConvNets exploit multiple modalities of data, the multi-modality information is often helpful in applications such as \cite{zhang2011fast,zhang2012fast,abeida2013iterative}.}) based methods have received tremendous attention in recent years, such as ``Two-Stream" \cite{simonyan2014two} and its derivatives \cite{wang2016two,wang2016temporal}. These methods are based on conventional 2D ConvNet, thus hardly requires video frame accumulation during prediction. Despite their lack of explicit 3D spatio-temporal ConvNet filters, they are still among the top performers on the UCF101~\cite{soomro2012ucf101} and HMDB51~\cite{Kuehne11} datasets. 

  Despite the successes, these ``Two-Stream" methods lack a holistic way of incorporating context information in video frames. Traditionally viewed as a source of noise, the background itself arguably offers valuable auxiliary information \cite{zhang2015multi, zhang2015auxiliary,zhang2014can} to recognize the characteristics of the human actions. A typical example is illustrated in Figure~\ref{bask_vs} (b). Even with the basketball player completely censored, the surrounding scene of the basketball court (especially the mid-air ball) provides substantial clues to recognize the ``playing basketball'' action. 
  %

\begin{figure*}[t]
    \centering
    \includegraphics[scale=0.22]{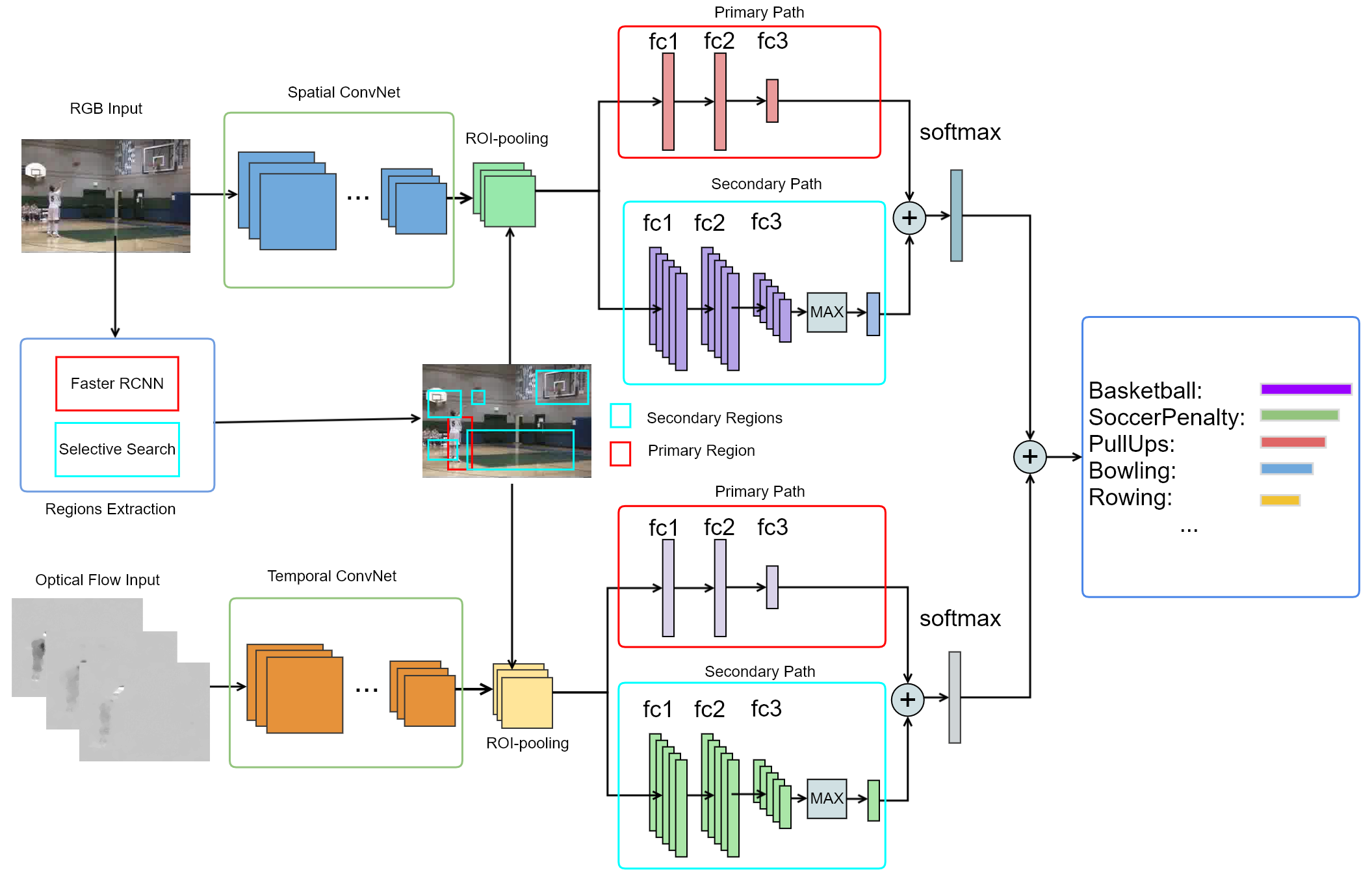}
    \caption{The architecture of WMR ConvNet. The RGB input is fed into a spatial ConvNet and a regions extraction pipeline separately. We use Faster R-CNN to extract primary region, i.e., the region of human bounding box(labeled as red bounding boxes) and Selective Search to extract secondary region proposals, i.e., the regions contain contexts and cues (labeled as cyan bounding box). After the ROI-pooling layer, primary region and selected secondary regions are fed into fully connected layers separately. Finally a softmax layer is set to fuse the scores from both primary region and secondary regions.　The stacked optical flow inputs reuse the regions of RGB input. After the output of each stream, a weighted sum is established to get final prediction.}
    \label{framework}
  \end{figure*}

  Based on the aforementioned intuition, the ``Weighted Multi-Region" (WMR) ConvNet is proposed to capture the context information from both the foreground and the background. Video frames are first fed into a fine-tuned Faster R-CNN~\cite{ren2015faster} network to localize the ``foreground'' with a bounding box, which is termed the primary region. Subsequently, multiple secondary regions per video frame are obtained with the Selective Search algorithm~\cite{uijlings2013selective} with varying intersection of Union (IoU) values with the primary region. Each testing image is fed into convolutional layers with the ROI-pooling~\cite{girshick2016region} over the extracted primary region and secondary regions, as illustrated in Figure~\ref{framework}. After the final stages of the RGB image stream and optical flow image stream, a fusion layer is added to account for the contributions from both streams, based on which the final predictions are made. 
  
  In the testing phase, the proposed WMR ConvNet processes frame-by-frame in the spatial ConvNet (RGB input) and 10-frame-after-10-frame in the temporal ConveNet (Optical flow input). Therefore, a maximum delay in processing is about 0.3 second (assuming 30 fps). On contrary, conventional action recognition methods typically require 3D convolution and LSTM-style temporal postprocessing, which could incur much higher latency. For example, video clip-based methods need to process multiple clips (each clip last about 0.5 seconds) before generating a prediction, which leads to latency of several seconds to tens of seconds.

  \noindent The major contributions of the paper are as follows.
  \begin{itemize}[noitemsep,leftmargin=*,topsep=0pt]
    \item A new WMR ConvNet with ``two-stream'' 2D ConvNet architecture suitable for low-latency online prediction. 
    \item An effective multi-region spatial sampling strategy that captures more informative spatial (RGB stream) and temporal (optical flow stream) contexts. 
    \item The proposed WMR ConvNet achieves the state-of-the-art recognition accuracy on both the UCF101 and HMDB51 dataset, among competing low-latency algorithms.
  \end{itemize}
  \section{Related Work}\label{sec:related}
  Action recognition has been extensively studied in past few years~\cite{wang2011action,wang2013action,cai2014effective,simonyan2014two,cai2015attribute,Tran_2015_ICCV,cai2016effective,carreira2017quo}. There are mainly two types of methods: conventional shallow feature based ones and those with deep neural network based ones. The improved Dense Trajectory (iDT)~\cite{wang2013action} method is the current state-of-the-art among the former type, in which the optical flow is utilized to estimate dense trajectories, followed by feature extractions around the trajectories. Inspired by the success of deep learning in tasks of image understanding, many researchers have attempted to incorporate deep architectures for action recognition. A major difference between image based tasks and video based tasks is the extra temporal information in videos, which is critical for action recognition. 

  One of the most successful architectures in action recognition is two-stream network~\cite{simonyan2014two}. In that paper, the authors use one CNN stream with the RGB frame as input to capture spatial information and another CNN stream with the stacked optical flow frames as input to capture temporal information. At the end of the last softmax layer, a score fusion method is established to merge spatio-temporal information. Besides these 2D ConvNet based approaches (with low-latency online prediction capability), the C3D architecture proposed in~\cite{Tran_2015_ICCV} is proved to be a good tool of extracting features for videos. In that paper, the authors extend traditional two dimensional convolution to three dimensional convolution, which captures spatio-temporal information adequately. However, these 3D ConvNet \cite{Tran_2015_ICCV,wang2016temporal} based methods require video frame accumulation during prediction, thus unfit for action recognition applications with low-latency prediction requirements.

  Since there are multiple cues in a still image, it's feasible to run action recognition on image level. R*CNN~\cite{gkioxari2015contextual} is an efficient work in action recognition of still images, in which R-CNN~\cite{girshick2014rich} is adapted to multiple proposal regions. Different from this method, our proposed method is deployed on video action recognition task, and the UCF101 and HMDB51 video datasets have no ground truth bounding boxes. 
  
  Another work that is similar to our method is Dynamic Network~\cite{bilen2016dynamic}, which summarizes the cues and information in video clips into a dynamic image, then feeds the dynamic image into a well-developed ConvNet to predict the label of the video. 
  \section{Proposed WMR ConvNet}\label{sec:attentionNet}
  The proposed WMR ConvNet is illustrated in Figure~\ref{framework}. Given an image $I$, we first feed it into a fine-tuned Faster R-CNN~\cite{ren2015faster} and Selective Search~\cite{uijlings2013selective} pipeline to obtain primary region (object proposal region with the highest confidence score) and secondary regions (other object candidate proposal regions other than the primary region), respectively. Then we input these annotations and $I$ to an adapted VGG-16 network~\cite{simonyan2014very}. An ROI Pooling layer is established to reuse the learnt feature maps in primary region and secondary regions. Then features of primary region and secondary regions are fed into three fully connected layers separately. In order to obtain the most informative secondary region for each action, we use max operation on the output of fully connected layers of the secondary region path. After obtaining maximum among secondary region scores, we add it to the primary score. Finally a softmax operation is established to transform scores into probabilities, which can be used to classify videos. Our full pipeline contains two paths of basic architecture: the one with RGB input and the one with optical flow input. At the end of the pipeline, a fusion scheme is established to predict the final classification probability. Each part is detailed below.
  
  \subsection{Region Proposal Generation}\label{sec:obtain_bbox}
  Since the two main benchmarks of human action recognition in video have no bounding box annotations,
  we come up with a solution to use Faster R-CNN to annotate them. We first annotate bounding boxes manually for $N_1$ and $N_2$ RGB frames in training set of UCF101 and HMDB51, respectively. We only label the bounding box that contains the person in foreground. We then fine-tune on Faster R-CNN using only two classes: background and person. We choose the box with the max score of person as the annotation of the frame. In this way we want to train a model that circles out action performers in videos automatically. Compared with using pre-trained model directly, our method can annotate person in videos more accurately.
  
  In order to obtain secondary regions, we first collect region proposals with Selective Search~\cite{uijlings2013selective} on RGB images. Then we choose the region proposals which have an overlap with the primary region of images. Formally, these regions can be defined as $U(r;I)$:
  \begin{equation}
  \label{eq:second_obtain}
  U(r;I)=\{s\in{S(I)}:\text{IoU}(s,r)\in{[l,u]}\},
  \end{equation}
  where $r$ is the primary region of the image $I$, $S(I)$ is the selective search result of $I$, $\text{IoU(s,r)}$ denotes the overlap ratio between region $s$ and $r$, $l$ and $u$ is the lower and upper boundary of overlap range, respectively.
  In the experiments we use the faster approximate implementation of Selective Search in Dlib (http://dlib.net/).
  When coming to optical flow images, they have no distinct secondary regions since in the flow images the still objects and background information are wiped away. We find that during a small temporal interval, the displacement of background and objects is usually tiny. Based on this observation, we reuse the union of bounding boxes of secondary regions from contiguous RGB frames as the secondary regions of the optical flow image. Formally, we have
  \begin{equation}
  U(r;O_{i})=U(r;I_{i})\cup U(r;I_{i+1}),
  \end{equation}
  where $O_{i}$ is the $i$th optical flow frame of a video $V$, $I_{i}$ and $I_{i+1}$ is the $i$th and $i+1$th RGB frames of $V$, respectively. For the primary region of an optical flow image, there are two ways to obtain it: 1) We get the rectangle region with the largest average magnitude as the bounding box. 2) We reuse the union of primary region of the two RGB frames where the optical flow is calculated. In a nutshell, by setting primary region and secondary regions reasonably for optical flow images, we extend the concept of cues and contexts to flow space. 
  
  \subsection{Learning Representations Using ConvNet}
  After obtaining bounding boxes, we feed the training image into a convolutional network (ConvNet) to learn a representation of the image. In this work we use VGG-16 architecture because of its good performance and relatively low time-cost during training and testing. In order to train the secondary regions effectively, we adopt ROI pooling in our work as in~\cite{girshick2016region}. Specifically, we replace the last pooling layer in VGG-16 with the ROI pooling layer, with the feature maps from previous layer and annotated region data as input. With it, each region generates a fixed-size feature map which can be fed into fully connected layers. Since the feature maps are all cropped from previous layer of ROI pooling (For example, the conv5-3 layer for VGG-16), no additional calculation is needed and this implementation is very efficient. 
  
  After the ROI pooling layer, we train the primary region and secondary regions separately to exploit potential cues. The primary region and secondary regions are separately fed to three fully connected layers to generate scores. Specifically, an extra max operation is carried out in the streams that process secondary regions. This max operation allows for the selection of the most informative secondary region for each action. Subsequently, this max score is fused with the primary region score with  weights of 0.6 for score from primary region and 0.4 for score from secondary regions. At the very end, a softmax layer processes the scores to obtain the predicted probability of each class.
  
  \subsection{Multi-task Learning}
  In order to learn the class of video and the accurate position of bounding box synchronously, we use the multi-task learning strategy to minimize the loss of the classification and loss of the bounding box regression at the same time. Given an image $I$, the final loss $L$ is the weighted sum of the loss of classification $L_{cls}$ and the loss of bounding box regression $L_{reg}$:
  \begin{equation}
  L = L_{cls}(I) + \alpha\times L_{reg}(r;U(r;I)),
  \end{equation}
  where $\alpha$ is the constant weight and $U(r;I)$ is the secondary regions of $I$. In this paper we set $\alpha$ to 0.3 according to ablation experiments. For more information, please refer to the Appendix of our manuscript.
  \subsection{Fusion of RGB Stream and Optical Flow Stream}
  In the subsections above, we discussed the basic framework of our method. Although we demonstrate our framework using RGB data, we claim here that the input of the framework is also applicable to optical flow data. Inspired by the popular Two-Stream architecture in action recognition task, we feed RGB data and optical flow data into our basic framework separately, then combine the outputs of these two networks, as shown in Figure~\ref{framework}. For each video, 1 frame of RGB data and 10 frames of optical flow data are fed into the basic architecture, respectively. After that, we merge extracted spatio-temporal features using a fusion scheme to get final prediction with weights of 0.4 and 0.6 for RGB stream and optical flow stream, respectively.
  \section{Experiments} \label{sec:experiments}
  We evaluate our method on UCF101~\cite{soomro2012ucf101} and HMDB51~\cite{Kuehne11}. UCF101 contains 13,320 video clips of 101 action classes collected from YouTube. HMDB51~\cite{Kuehne11} contains 6766 video clips from movies and YouTube, annotated into 51 action classes. For each of those two datasets, approximately two thousand video frames are manually labeled with bounding boxes of persons to fine-tune the faster R-CNN network. The output bounding box with the highest R-CNN confidence score within a video frame is treated as the ``foreground'' in our framework.
  
  \subsection{ConvNet Configuration}
  Our implementation is built on R*CNN~\cite{gkioxari2015contextual}, in which the VGG-16 network is used. Instead of using max pooling, an ROI pooling layer is established as pool5 to classify the multiple ROI regions. After that layer, we set up two paths of fully connected layers to learn presentations of the primary region and secondary regions separately. 
  Basically, we train our architecture with stochastic gradient descent (SGD) using back propagation (BP). We fine-tune a VGG-16 model pretrained on ImageNet~\cite{deng2009imagenet}. We set initial learning rate to 0.0001 and divide it by 10 after every 50000 iterations. We set the dropout ratio to 0.6 according to ablation experiments offered in Appendix. We set the batch size to 256 and use 2 images per batch. In total, our model is trained with 200K iterations. 
  
  For RGB data, we randomly choose 1 frame from each training video as input according to the settings in~\cite{simonyan2014two}. For optical flow data, we choose 10 frames from each video. When testing, we sample 25 frames of RGB data and flow data for each testing video to obtain the final result. 

  \subsection{Evaluation}
\begin{figure}[t]
    \centering
    \includegraphics[scale=0.16]{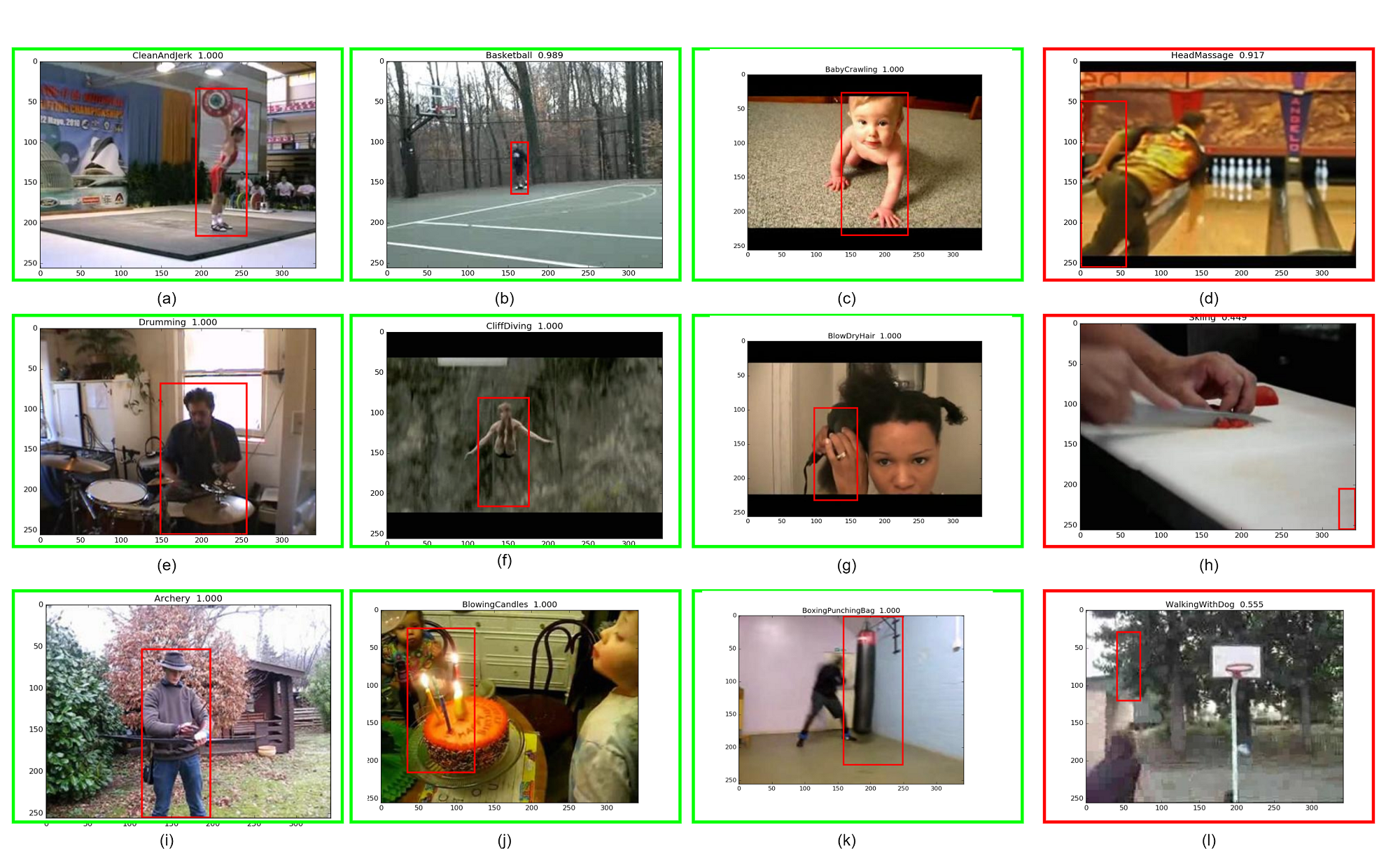}
    \caption{The human detection results. Note the first three columns of each row, colored in green, are rightly labeled data, the fourth column of each row, colored in red, is wrongly labeled data. At the top of each subplot, there is a testing class name and its confidence. The true label of these images are (from left to right, top to bottom): CleanAnjerk, Basketball, BabyCrawling, Bowling, Drumming, CliffDiving, BlowDryHair, CuttingInKitchen, Archery, BlowingCandles, BoxingPunchingBag and Basketball.}
    \label{good_bad_example}
  \end{figure}

  In this subsection we conduct the human body detection and evaluate its impact on the final result. We label $N_1=2525$ RGB images from UCF101, $N_2=2177$ RGB images from HMDB51 to run transfer learning on Faster R-CNN. After fine-tuning the Faster R-CNN network, we input a testing image to the network and select the bounding box with the highest score of person category as the primary region, which is shown with a red box in each sub figure in Figure~\ref{good_bad_example}. In each row, the first three samples are correctly annotated (with a green box outside) and the fourth sample is wrongly annotated (with a red box outside). 
  
  From Figure~\ref{good_bad_example} we find that the accuracy of annotation affects the final results heavily. For example, in $(d)$, a hand of a person playing the bowling (the true label of this frame) is detected and its pose is similar to pose in ``HeadMessage" videos the ConvNet has seen, thus the testing label is set to it. In $(l)$, a small part of a green tree is detected, which is similar to green grasses in ``WalkingWithDogs" videos. In $(h)$, a white chopping board similar to the snowfield is detected so the ConvNet classifies this frame to ``Skiing". These examples show that the region detected as the primary region is vital for the action recognition task.
  
  In order to obtain secondary regions, we run Selective Search firstly to get all region proposals. Then select regions among them using Equation~\eqref{eq:second_obtain} by setting $l$ to 0.1 and $u$ to 0.9.

  As discussed in section~\ref{sec:obtain_bbox}, we use two approaches to get the primary regions of optical flow images. The results are shown in Table~\ref{table:obtain_bbox}. The approach reusing bounding boxes of RGB image is better, which indicates that the cues in RGB data are more useful than magnitude of optical flow. 
  \setlength{\tabcolsep}{6pt}
  \begin{table}[t]
    \centering
    \caption{Comparison of optical results using different primary region annotation methods. In 1), a rectangle with max average magnitude is selected as the primary region; in 2), the bounding box of corresponding RGB frame is selected as the primary region. We conduct experiments on $u$ channel and $v$ channel separately. Please check the detailed description in section~\ref{sec:obtain_bbox}.}
    \label{table:obtain_bbox}
    \begin{tabular}{l|c|c}
      \hline
      Method 	& Accuracy ($u$ channel)	& Accuracy ($v$ channel)\\
      \hline												
      1)		& 75.3\% 				& 75.3\% \\		
      2)		& \textbf{76.2\%} 	& \textbf{77.4\%} \\		
      \hline
    \end{tabular}
  \end{table}
  
  In addition to weighted sum, we explore another classifier to fuse features of the two-path networks. We propose two alternatives here: 1) For features extracted from fully connected layers, we concatenate features from primary region and secondary regions. After that, we use a linear SVM to classify them. 2) For scores obtained after softmax layer, we use either weighted sum or a linear SVM.
  In Table~\ref{table:fusion_ps}, we find that results with weighted sum operation are the best. The feature of fc1 layer and fc2 layer are worse than score layer features and weighted sum outperforms SVM on score layer features.
  \setlength{\tabcolsep}{34pt}
  \begin{table}[t]
    \centering
    \caption{Comparison of different fusion method on RGB data of UCF101 split-1. For both SVM and weighted sum algorithm, we extract target features from training dataset in both primary region network and secondary regions network and then concatenate them to train a classifier. Then we do the same thing on the testing image to predict the class.}
    \label{table:fusion_ps}
    \begin{tabular}{l|c}
      \hline
      Fusion method							& Accuracy \\		
      \hline												
      fc1 + SVM								& 61.1\% \\
      fc2 + SVM								& 66.5\% \\
      score + SVM								& 73.7\% \\
      score + weighted sum 						& \textbf{75.9}\% \\
      \hline																	
    \end{tabular}
  \end{table}
  %
  %
  %
  
  Here we compare the proposed ablated WMR ConvNet with the dynamic image networks (MDI, MDI + static RGB). For a fair comparison, only RGB data is used as inputs to the ablated WMR ConvNet.
  Table~\ref{table:compare_din} shows the results of our method and dynamic image networks. We find that our method outperforms the dynamic image networks. This indicates the necessity of encoding the cues in video for our method. 
  \setlength{\tabcolsep}{14pt}
  \begin{table}[h]
  \centering
      \caption{Comparison with the Dynamic Network~\cite{bilen2016dynamic}. In order to compare fairly, we only use the RGB data as input to our method.}
      \label{table:compare_din}
      \begin{tabular}{l|c}
        \hline
        Method 						 		& UCF101 	\\
        \hline
        static RGB					 		& 70.1\% 	\\
        MDI				 		& 70.9\% 	\\
        MDI + static RGB 		& 76.9\% 	\\
        ablated WMR ConvNet (RGB only)                	& \textbf{78.8\%} \\
        \hline
      \end{tabular}
  \end{table}
  
  \setlength{\tabcolsep}{16pt}
  \begin{table}[t]
    \centering
    \caption{Comparison of proposed WMR ConvNet with competing methods on UCF101 and HMDB51 datasets. }
    \label{table:all_result}
    \begin{tabular}{l|c|c}
      \hline
      Method &  UCF101 & HMDB51 \\
      \hline
      iDT~\cite{wang2013action} 							& 85.9\%  & 57.2\%\\
      Slow Fusion~\cite{karpathy2014large} 		 		& 65.4\% & - \\
      Two Stream~\cite{simonyan2014two}				 	& \textbf{88.0\%} &  59.4\%\\
      C3D~\cite{Tran_2015_ICCV}					  		& 85.2\% &  -\\
      WMR ConvNet											& 85.7\% & \textbf{66.7}\%\\
      \hline
    \end{tabular}
  \end{table}

  From Table~\ref{table:all_result}, we find that our method is comparable with iDT and C3D on UCF101 dataset, and outperforms other methods on HMDB51 dataset. Note that C3D~\cite{Tran_2015_ICCV} requires video frame accumulation during prediction, thus it is not directly compatible with low-latency online prediction requirements. This indicates that our proposed WMR ConvNet is effective and spatio-temporal cues and contexts are crucial for human action recognition in videos.
  \setlength{\tabcolsep}{26pt}
  \begin{table}[h]
    \centering
    \caption{Results of spatial stream ConvNet (RGB input only) on UCF101.}
    \label{table:compare_spatial}
    \begin{tabular}{l|l}
      \hline
      Method 											& Accuracy	\\
      \hline
      Two-Stream~\cite{simonyan2014two}				& 73.0\% 	\\
      DIN~\cite{bilen2016dynamic}						& 76.9\% 	\\
      SR-CNN~\cite{wang2016two}						& 78.3\% 	\\
      Ablated WMR ConvNet              				& \textbf{78.8\%} \\
      \hline
    \end{tabular}
  \end{table}
  Here we compare our results on spatial stream ConvNet with other published results. Table~\ref{table:compare_spatial} shows that our method outperforms the compared methods in spatial domain. This superior performance demonstrates the effectiveness of exploiting multi-region based cues in spatial space and indicates that our method is more suitable for low-latency usage than compared methods.
  %
  %
  \section{Conclusions}\label{sec:conclusions}
  In this paper, we propose the WMR ConvNet for action recognition with low-latency online prediction requirements. An efficient multi-region spatial sampling strategy is proposed to explicitly capture both human related information (e.g., poses) and context information, via primary region and secondary regions, respectively. Such regions are also shared by the optical flow pipeline, which offers additional temporal information. Even with the optical flow pipeline removed, the ablated WMR ConvNet achieves a high recognition accuracy of 78.8\% on the UCF101 dataset. The complete WMR ConvNet outperforms all competing 2D ConvNet based algorithms (with low-latency online prediction) and even a 3D ConvNet based C3D algorithm on the UCF101 dataset. 
  
  \bibliographystyle{IEEEbib}
  \bibliography{WMR}

  \begin{appendices}
  \section{Sensitivity Analysis on Network Parameters}
Here we will show results on split-1 of UCF101~\cite{soomro2012ucf101} using only RGB data with different dropout ratios.
\subsection{Dropout}
Dropout is an important technique in deep convolutional network, which can be used to avoid overfitting. In this part, we conduct experiments on different dropout ratios during the training phase of the network. 
\setlength{\tabcolsep}{36pt}
\begin{table}[h]
\centering
		\caption{Comparison of results using different dropout ratios on RGB data of UCF101 split-1. }
		\label{table:dropout}
		\begin{tabular}{c|c}
			\hline
			Dropout ratio 							& Accuracy \\		
			\hline												
			0.5										& 75.6\% \\
			0.6										& \textbf{76.7}\% \\
			0.9										& 37.7\% \\
			\hline		
		\end{tabular}
\end{table}

As shown in Table~\ref{table:dropout}, we achieve the best result when setting dropout ratio to 0.6. Surprisingly, we only get humble 37.7\% when setting dropout to 0.9 even after 200K iterations. Since 0.9 dropout ratio achieves quite good result on two-stream architecture (See Table 1 (a) in~\cite{simonyan2014two}), it indicates that multi-task architecture is more difficult to learn and is less likely to overfit. 

\end{appendices}
  \end{document}